\documentclass[10pt, a4paper]{article}

\usepackage{arabtex}
\usepackage{pdfpages}
\usepackage{subcaption}
\usepackage{utf8}
\setcode{utf8}
\def\endabstract{\egroup}

\usepackage{xfakebold}
\newcommand{\fbseries}{\unskip\setBold\aftergroup\unsetBold\aftergroup\ignorespaces}

\usepackage{amssymb}
\usepackage{booktabs} 
\usepackage{array}
\usepackage{colortbl}
\usepackage{xcolor}
\usepackage{makecell}
\usepackage{graphicx}
\usepackage{multirow}
\usepackage{enumitem}
\usepackage{footmisc}
\usepackage{amssymb}

 \newcolumntype{C}[1]{>{\centering\arraybackslash}m{#1}}

\usepackage[final]{lrec-coling2024} 

\title{Leveraging Corpus Metadata to Detect Template-based Translation: An Exploratory Case Study of the Egyptian Arabic Wikipedia Edition}

\name{\hspace{-4pt}Saied Alshahrani$^1$ Hesham Haroon$^2$ Ali Elfilali$^3$ Mariama Njie$^4$ Jeanna Matthews$^1$} 
\address{\hspace{-5pt}$^1$Clarkson University, USA $^2$Sesame Labs, Egypt $^3$Cadi Ayyad University, Morocco $^4$M\&T Bank, USA\\
\hspace{-6pt}\{\href{mailto:saied@clarkson.edu}{\color{black}{saied}}, \href{mailto:jnm@clarkson.edu}{\color{black}{jnm}}\}@clarkson.edu, hesham@smsm.ai, a.elfilali9805@uca.ac.ma, mnjie@mtb.com}

\abstract{
Wikipedia articles (content pages) are commonly used corpora in Natural Language Processing (NLP) research, especially in low-resource languages other than English. Yet, a few research studies have studied the three Arabic Wikipedia editions, Arabic Wikipedia (AR), Egyptian Arabic Wikipedia (ARZ), and Moroccan Arabic Wikipedia (ARY), and documented issues in the Egyptian Arabic Wikipedia edition regarding the massive automatic creation of its articles using template-based translation from English to Arabic without human involvement, overwhelming the Egyptian Arabic Wikipedia with articles that do not only have low-quality content but also with articles that do not represent the Egyptian people, their culture, and their dialect. In this paper, we aim to mitigate the problem of template translation that occurred in the Egyptian Arabic Wikipedia by identifying these template-translated articles and their characteristics through exploratory analysis and building automatic detection systems. We first explore the content of the three Arabic Wikipedia editions in terms of density, quality, and human contributions and utilize the resulting insights to build multivariate machine learning classifiers leveraging articles’ metadata to detect the template-translated articles automatically. We then publicly deploy and host the best-performing classifier, XGBoost, as an online application called \textsc{Egyptian Wikipedia Scanner}$^\clubsuit$ and release the extracted, filtered, and labeled datasets to the research community to benefit from our datasets and the online, web-based detection system.
 \\ \newline  
 }

\newcommand\blfootnote[1]{%
  \begingroup
  \renewcommand\thefootnote{}\footnote{#1}%
  \addtocounter{footnote}{-1}%
  \endgroup
}

\begin{document}

\maketitleabstract

\section{Introduction}
\label{sec:1}
\blfootnote{\hspace{-5pt}$^\spadesuit$Accepted \hspace{1.5pt}at \hspace{1.5pt}the \hspace{1.5pt}LREC-COLING \hspace{1.5pt}2024: \hspace{1.5pt}OSACT6.}
\blfootnote{\hspace{-5pt}$^\clubsuit$\href{https://huggingface.co/spaces/SaiedAlshahrani/Egyptian-Wikipedia-Scanner}{https://hf.co/spaces/Egyptian-Wikipedia-Scanner}.}
Wikipedia articles are widely used as pre-training datasets for many Natural Language Processing (NLP) tasks like language modeling (language models) and word representation (word embedding models) tasks, especially for low-resource languages like Arabic, due to its large collection of multilingual content and its vast array of metadata that can be quantified and compared \citep{112}. However, not all Wikipedia articles are organically produced by native speakers of those languages; while humans have naturally generated some articles in those languages, many others have been automatically generated using bots or automatically translated from high-resourced languages like English without human revision using off-the-shelf automatic translation tools like Google Translate\footnote{Google Translate: \href{https://translate.google.com}{https://translate.google.com}.} \citep{4, 2, 1, 11, 3, 5, 8}.

A few researchers have addressed this issue and highlighted its implications for NLP systems and tasks. For example,  \citet{11} have studied the three Arabic Wikipedia editions, Arabic Wikipedia (AR), Egyptian Arabic Wikipedia (ARZ), and Moroccan Arabic Wikipedia (ARY), and documented issues in the Egyptian Wikipedia with automatic creation/generation and translation of content pages from English without human supervision. They stressed that these issues could substantially affect the performance and accuracy of Large Language Models (LLMs) trained from these corpora, producing models that lack the cultural richness and meaningful representation of native speakers. In another research work by the same authors, they investigated the performance implications of using inorganic, unrepresentative corpora, mainly generated through automated techniques such as bot generation or automated template-based translation, to train a few masked language models and word embedding models. They found that models trained on bot-generated or template-translated articles underperformed the models trained on human-generated articles and underscored that, for good NLP performance, researchers need both large and organic corpora \citep{13}.

In this paper, we solely focus on the problem of template translation that took place in the Egyptian Arabic Wikipedia edition, where a few registered users employed simple templates to translate more than one million content pages (articles) from English to Arabic using Google Translate, all without translation error checking or culture misrepresentation verification, disregarding the consequences of using such poor articles \citep{1, 7, 11, 111, 9, 10}. We first explore the three Arabic Wikipedia editions’ content in terms of density, quality, and human contributions, highlighting how the template-based translation occurred on the Egyptian Wikipedia produces unrepresentative content. We second, attempt to build powerful multivariate machine learning classifiers leveraging corpus/articles’ metadata to detect the template-translated articles automatically. We then deploy and publicly host the best-performing classifier, XGBoost, so researchers, practitioners, and other users can benefit from our online, web-based detection system. We lastly argue that practices such as template translations could not only impact the performance of models trained on these template-translated articles but also could misrepresent the native speakers and their culture and do not echo their views, beliefs, opinions, or perspectives.

\section{Exploratory Analysis}
\label{sec:2}
We explore, in the following subsections, the three Arabic Wikipedia editions, Arabic Wikipedia (AR), Egyptian Arabic Wikipedia (ARZ), and Moroccan Arabic Wikipedia (ARY), regarding their articles’ content in terms of density, quality, and human contributions.

\subsection{Analysis Setup}
\label{sec:2.1}
We follow the same methodology \citet{13} used to quantify the bot-generated articles, but we, here, utilize the Wikimedia \texttt{XTools} API\footnote{XTools API: \href{https://www.mediawiki.org/wiki/XTools}{https://www.mediawiki.org/wiki/XTools}.} to collect all Arabic Wikipedia editions’ articles’ metadata; specifically, we collect the total edits, total editors, top editors, total bytes, total characters, total words, creator name, and creation date for each article. We use the complete Wikipedia dumps of each Arabic Wikipedia edition, downloaded on the 1st of January 2024 \citep{14} and processed using the \texttt{Gensim} Python library  \citep{15}. We also use Wikipedia’s “List Users” service\footnote{\href{https://en.wikipedia.org/wiki/Special:ListUsers}{https://\{WIKI\}.wikipedia.org/wiki/Special:ListUsers}.} to retrieve the full list of bots in each Arabic Wikipedia edition to measure the bot and human contributions to each article.

\subsection{Shallow Content}
\label{sec:2.2}
We, in this subsection, study the density of the content of the three Arabic Wikipedia editions, highlighting general statistics and token/character length distributions per Arabic Wikipedia edition.

\subsubsection{Summary Statistics}
\label{sec:2.1.1}
We shed light on a few general statistics of the three Arabic Wikipedia editions regarding their total articles, total extracted articles, corpus size, total bytes, total characters, and total tokens, highlighting the minimum, maximum, and mean values of the three articles’ metadata: total bytes, total characters, and total tokens.\footnote{We use the Wikimedia Statistics service, \href{https://stats.wikimedia.org}{https://stats.wikimedia.org}, to retrieve the total articles (content pages) for each Arabic Wikipedia edition, whereas the other statistics are generated from the extracted articles from each Arabic Wikipedia edition.} From Table \ref{tab:1}, it is notable that the Egyptian Arabic Wikipedia has a greater number of total articles than the Arabic Wikipedia (which is generally believed to be more organically generated), with almost 400K articles, yet as we will discuss later in Table \ref{tab:3}, this number of total articles does not reflect true measurements of organically generated contributions since all three Arabic Wikipedia editions include substantial bot generation and template translation activities \citep{1, 11, 12}. We employ the \texttt{Gensim} Python library to parse and extract the textual content (articles) from each Wikipedia dump file. However, since the library discards any articles with less than 50 tokens/words, all three Arabic Wikipedia editions lost many articles. For example, the Egyptian Wikipedia lost nearly 741K (46\%) of its articles, whereas the Moroccan Wikipedia and the Arabic Wikipedia lost 2.9K (30\%) and 346K (28\%) of their articles, respectively. This loss of articles exhibits how the Egyptian Arabic Wikipedia contains almost half of its total articles under 50 tokens per article, indicating that it has more limited and shallow content and reflecting the template translation that occurred on its articles.

\begin{table*}[!htp]
\centering
\tiny\renewcommand{\arraystretch}{1}
\resizebox{\textwidth}{!} {
\begin{tabular}{|c|c|c|c|c|c|c|c|c|c|c|c|c|} 
\hline
\multirow{2}{*}{\textbf{Wikipedia}} & \multirow{2}{*}{\makecell {\textbf{Total}\\ \textbf{Articles}}} & \multirow{2}{*}{\makecell {\textbf{Extracted}\\ \textbf{Articles}}} & \multirow{2}{*}{\makecell {\textbf{Corpus}\\ \textbf{Size}}} & \multicolumn{3}{c|}{\textbf{Total Bytes}} & \multicolumn{3}{c|}{\textbf{Total Characters}} & \multicolumn{3}{c|}{\textbf{Total Tokens}} \\
\cline{5-13}
   & & & & {\textbf{Min}} & {\textbf{ Max }} &{\textbf{ Mean }}& {\textbf{ Min }} & {\textbf{Max}}& {\textbf{Mean}} & {\textbf{ Min$^\bigstar$ }}  & {\textbf{ Max}} & {\textbf{ Mean }}\\
\hline
\multirow{2}{*}{\makecell{\textbf{Arabic (AR)}}} & \multirow{2}{*}{1,226,784} & \multirow{2}{*}{880,334} & \multirow{2}{*}{2.6GB} & \multicolumn{3}{c|}{6,424,572,842} & \multicolumn{3}{c|}{1,564,243,778} & \multicolumn{3}{c|}{264,761,062} \\
\cline{5-13}
   & & & & \hfil 488 & 1,419,547 & 7,297 & 200 & 334,464& 1,776 & 50 & 56,395 & 300 \\
\hline
\multirow{2}{*}{\makecell{\textbf{Egyptian (ARZ)}}} & \multirow{2}{*}{1,621,745} & \multirow{2}{*}{736,158} & \multirow{2}{*}{766MB} & \multicolumn{3}{c|}{1,525,938,072} & \multicolumn{3}{c|} {449,449,693} & \multicolumn{3}{c|}{74,277,188} \\
\cline{5-13}
   & & & & 515 & 1,217,036 & 2,072 & 233 & 399,641 & 610 & 50 & 74,009 & 100 \\
\hline
\multirow{2}{*}{\makecell{\textbf{Moroccan (ARY)}}} & \multirow{2}{*}{9,659} & \multirow{2}{*}{6,754} & \multirow{2}{*}{11MB} & \multicolumn{3}{c|}{25,109,824} & \multicolumn{3}{c|}{6,802,694} & \multicolumn{3}{c|}{1,153,946} \\
\cline{5-13}
   & & & & 646 & 105,009 & 3,717 & 248 & 32,853 & 1,007 & 50  & 5,635 & 170\\
\hline
\end{tabular}}
\caption{\label{tab:1}General statistics of the three Arabic Wikipedia editions in terms of total articles, total extracted articles, corpus/articles size, total bytes, total characters, and total tokens. $^\bigstar$As a result of the \texttt{Gensim} Python library discarding articles with tokens/words less than 50, all minimum tokens of articles are 50.}
\vspace{-3pt}
\end{table*}

\subsubsection{Token/Character Length Distribution}
\label{sec:2.2.2}
\begin{figure*}[!htp]
    \centering
    \includegraphics[width=\linewidth]{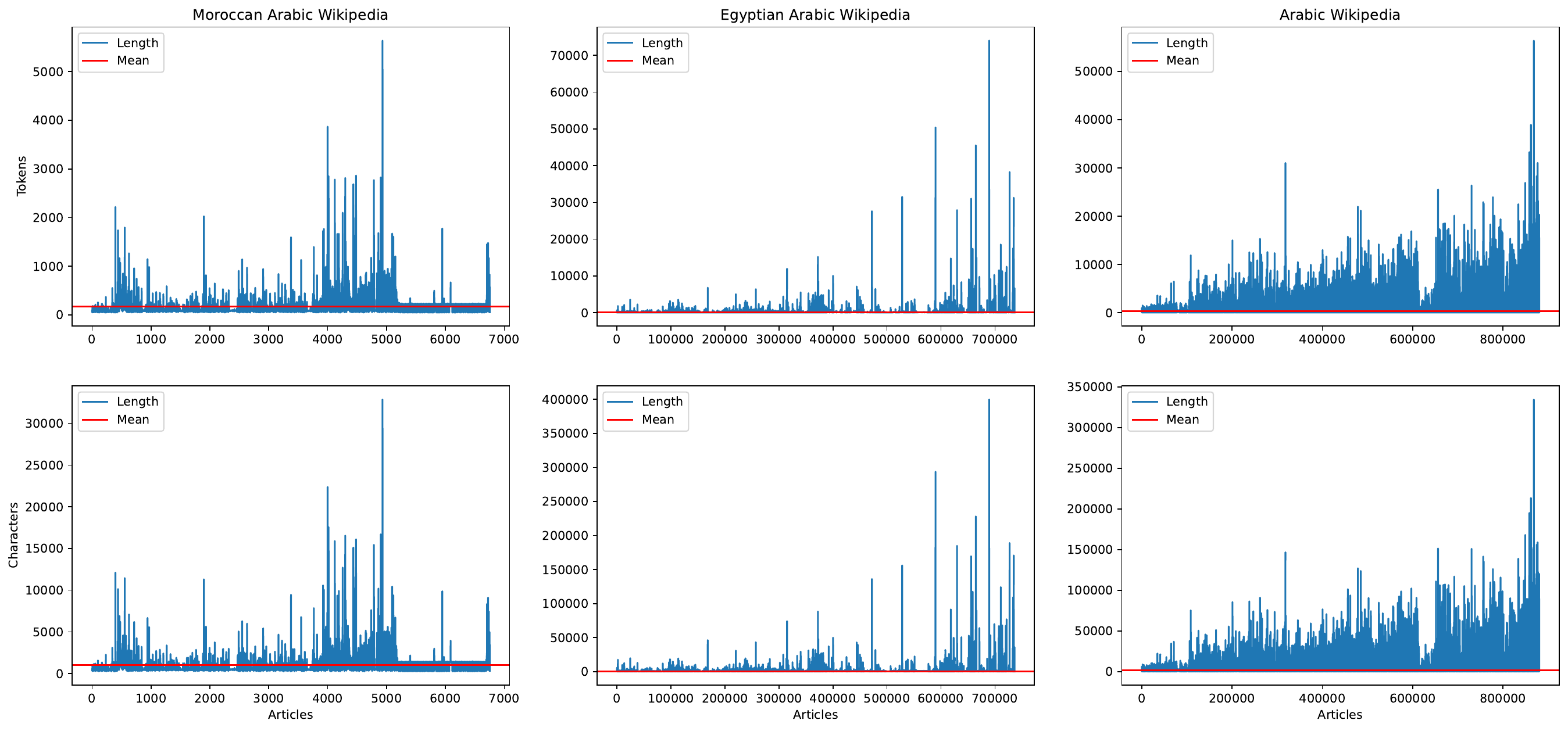}
    \caption{Visualizations of tokens and characters per article for each Arabic Wikipedia edition, displaying the total tokens and characters on the y-axes and articles on the x-axes, with plotting the mean lines.}
    \label{fig:1}
    \vspace{-3pt}
\end{figure*}

We visualize, in Figure \ref{fig:1}, the token and character distributions for each Arabic Wikipedia edition by plotting the tokens per article and characters per article with the mean lines for each Arabic Wikipedia edition. We observe that the Egyptian Wikipedia length distributions (token and character) are less dense than the Arabic Wikipedia and Moroccan Wikipedia, and a notable number of articles in the Egyptian Wikipedia are below the mean line/threshold, exhibiting that the Egyptian Wikipedia has unusually smaller and shorter articles than other Arabic Wikipedia editions. Surely, the Egyptian Wikipedia has more articles than the other Arabic Wikipedia editions, but it does have the lowest mean values of the total of characters and total of tokens/words, 610 and 100, respectively, compared to the mean values of the Arabic Wikipedia and the Moroccan Wikipedia, as shown in Table \ref{tab:1}. These observations signal that the template translation that happened on its articles does not produce rich, dense, and long content but only produces poor, limited, and shallow content.

\subsection{Poor Quality Content}
\label{sec:2.3}
We study the quality of the Arabic Wikipedia editions’ content regarding lexical richness and diversity and the most common and duplicate n-grams.

\subsubsection{Lexical Richness/Diversity}
\label{sec:2.3.1}
We use the terms lexical richness and lexical diversity equivalently and interchangeably in this study, as \citet{16} suggested. To measure the lexical richness and diversity, we first compute the total tokens and unique tokens per Arabic Wikipedia edition, and second, we utilize three simple but widely used lexical richness metrics: Type-Token Ratio (TTR) \citep{17, 18}, Root Type-Token Ratio (RTTR) \citep{19, 20}, and Corrected Type-Token Ratio (CTTR) \citep{21}. Yet, as many have emphasized, like \citet{22}, we find that these metrics are not often precise and sometimes erroneous and do not reflect the true lexical richness and diversity of a corpus. For example, we observe that the TTRs of Arabic Wikipedia and Egyptian Wikipedia are identical, and the  RTTRs and CTTRs of Egyptian Wikipedia and Moroccan Wikipedia are similar, despite the massive difference between the Arabic Wikipedia editions’ corpora in terms of the lexicon size and vocabulary size, as shown in Table \ref{tab:2}. Therefore, we adopt an advanced metric to measure the lexical richness and diversity called ‘Measure of Textual Lexical Diversity (MTLD)’, introduced by \citet{23}. We utilize the \texttt{LexicalRichness} Python library’s implementation of the MTLD metric with a default factor size of 0.720 \citep{24}. We find that the results are consistent with the other metrics, as reported in Table \ref{tab:2}, in that the Moroccan Wikipedia has the best lexical richness and diversity among the three Arabic Wikipedia editions, where the Arabic Wikipedia comes second, and Egyptian Wikipedia comes in last, documenting the Egyptian Arabic Wikipedia corpus is not lexically rich and diverse, which we attribute to the template-based translation took place on its articles (content pages).

\begin{table*}[!htp]
\centering
\tiny\renewcommand{\arraystretch}{1}
\resizebox{\textwidth}{!} {
\begin{tabular}{|c|c|c|c|c|c|c|} 
\hline
{\textbf{Wikipedia}} & {\makecell {\textbf{Total}\\ \textbf{Tokens}}} & {\makecell {\textbf{Unique}\\ \textbf{Tokens}}} & {\makecell {\textbf{Type-Token}\\ \textbf{Ratio (TTR)}}} & {\makecell {\textbf{Root Type}\\ \textbf{Token Ratio }\textbf{(RTTR)}}} & {\makecell {\textbf{Corrected Type}\\ \textbf{Token Ratio } \textbf{(CTTR)}}} & {\makecell {\textbf{Measure of Textual}\\ \textbf{ Lexical Diversity} \textbf{(MTLD)}}} \\
\hline
{\textbf{Arabic (AR)}} & 264,777,392 & 2,867,782 & 0.010 & 176.24 & 124.62 & 71.20 \\
\hline
{\textbf{Egyptian (ARZ)}} & 74,278,320 & 759,519 & 0.010 & 88.12 & 62.31 & 45.69 \\
\hline
{\textbf{Moroccan (ARY)}} & 1,154,058 & 94,827 & 0.082 & 88.27 & 62.41 & 89.77\\
\hline
\end{tabular}}
\caption{\label{tab:2}Calculations of four lexical richness and diversity metrics, TTR, RTTR, CTTR, and MTLD, accompanied with total tokens (lexicon) and unique tokens (vocabulary) for each Arabic Wikipedia edition.}
\end{table*}

\subsubsection{Most Common/Duplicate N-Grams}
\label{sec:2.3.2}
We generate n-grams from each Arabic Wikipedia corpus, where n=\{1, 2, 3, 5, 10, 50\}, to highlight the common and duplicate n-grams. We hypothesize that the higher the count of n-grams in an Arabic Wikipedia corpus, especially when n=\{5, 10, 50\}, the more we can detect templates used in the template translation activities in the Arabic Wikipedia editions, specifically in the Egyptian Wikipedia. We notice that n-grams in the Egyptian Wikipedia have very large counts compared to the Arabic and Moroccan Wikipedia editions, as shown in Tables \ref{tab:9} and \ref{tab:10} in Appendix \ref{app:a}.\footnote{We further analyze the 5-grams and 10-grams of each Arabic Wikipedia edition in Appendix \ref{app:a}.} In Figure \ref{fig:2}, we visualize the log values of the top K=1 counts of common and duplicate n-grams generated from each Arabic Wikipedia corpus, where n=\{1, 2, 3, …, 50\}, and we observe that all the n-grams in all the Arabic Wikipedia editions exhibit exponential decay, drastically (like Arabic Wikipedia) or gradually (like Egyptian Wikipedia and Moroccan Wikipedia). Yet, the large counts of Egyptian Wikipedia’s n-grams when n>=\{5\} do not decline exponentially but linearly, suggesting that there is an anomaly in the Egyptian Wikipedia corpus, where the n-grams of the normally generated corpus by humans usually factorially decreases, as the n value increases. We believe the template-based translation on the Egyptian Wikipedia creates such an anomaly, as many parts/grams/phrases of templates used in the translation are duplicated repeatedly in its corpus.

\begin{figure}[!htp]
    \centering
    \includegraphics[width=\linewidth]{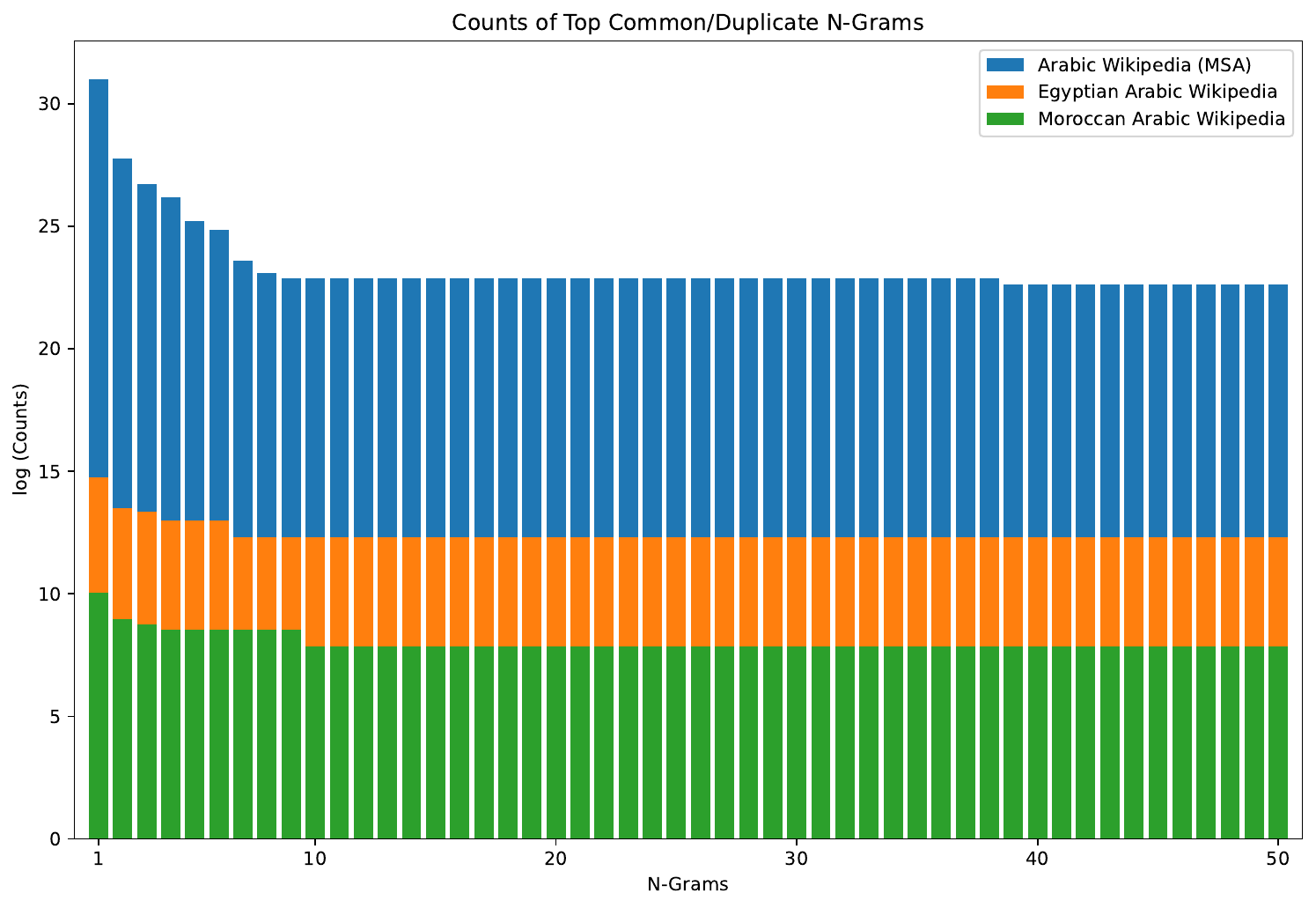}
    \caption{Counts of top common/duplicate n-grams of each Arabic Wikipedia edition; log values/counts are only for top K=1 common/duplicate n-grams.}
    \label{fig:2}
\end{figure}


\subsection{Misleading Human Involvement}
\label{sec:2.4}
We shed light on the human involvement across the three Arabic Wikipedia editions, specifically the type of page creators and editors, debating how the template translation activities could produce misleading metadata regarding human involvement.

\subsubsection{List of Contributors}
\label{sec:2.4.1}
We collect all the page creators for each article in the Arabic Wikipedia editions, count the number of their contributions (article creations), and categorize them into bots and humans. As shown in Table \ref{tab:3}, it is clear that the Arabic Wikipedia and Moroccan Wikipedia suffer from mass auto-creation of articles by bots, especially by the ‘JarBot’, which has created nearly 260K articles (29.31\%) in the Arabic Wikipedia, and the ‘DarijaBot’, which has created nearly 3.2K articles (34\%) in the Moroccan Wikipedia.\footnote{These two bots, ‘JarBot’ and ‘DarijaBot’, have approval from Wikimedia to operate on the Arabic Wikipedia and the Moroccan Wikipedia \citep{114, 115}.} However, the worst of all is the unguided, unreviewed, unsupervised template translation of articles from English in the Egyptian Wikipedia by registered users, largely by two registered users, ‘HitomiAkane’ and ‘Al-Dandoon’, who have created more than 1.4M articles (88.57\%) and 113K articles (6.99\%), respectively.\footnote{These two registered users were local admins of the Egyptian Arabic Wikipedia edition until their permissions were revoked in May 2020 by the Stewards, the global admins of the Wikipedia project, for their abuse of admin permissions and their massive unsupervised and unauthorized creation of articles \citep{51}.}

\begin{table*}[h!]
\centering
\tiny\renewcommand{\arraystretch}{1}
\resizebox{\textwidth}{!} {
\begin{tabular}{|c|c|c|c|c|c|c|} 
\hline 
\multicolumn{2}{|c|}{\textbf{Wikipedia $\backslash$ Rank (percentage)}} & {\makecell {\textbf{1st} \textbf{(\%)}}}& {\makecell {\textbf{2nd} \textbf{(\%)}}}  &  {\makecell {\textbf{3rd} \textbf{(\%)}}} &  {\makecell {\textbf{4th} \textbf{(\%)}}} &   {\makecell {\textbf{5th} \textbf{(\%)}}} \\
\hline
\multirow{3}{*}{\textbf{Arabic (AR)}} & {\textbf{Name}} & JarBot & Mr. Ibrahem & \<جار الله> & ElphiBot & Majed \\ \cline{2-7}
 & {\textbf{Count}} & \makecell{359,677 (29.31\%)}  & \makecell{52,222 (4.25\%)} & \makecell{43,691 (3.56\%)} & \makecell{42,669 (3.47\%)} & \makecell{26,228 (2.13\%)} \\ \cline{2-7}
 & {\textbf{Type}} & Bot & Human & Human & Bot & Human \\
\hline
\multirow{3}{*}{\textbf{Egyptian (ARZ)}} & {\textbf{Name}} & HitomiAkane & Al-Dandoon & Raafat & Ghaly & 10\<حمدى>\\ \cline{2-7}
 & {\textbf{Count}} & \makecell{1,436,430 (88.57\%)} & \makecell{113,468 (6.99\% )}& \makecell{18,334 (1.13\%)} & \makecell{7,212 (0.44\%)}& \makecell{2,720 (0.16\%)}  \\ \cline{2-7}
 & {\textbf{Type}} & Human & Human & Human & Human & Human \\
\hline
\multirow{3}{*}{\textbf{Moroccan (ARY)}} & {\textbf{Name}} & DarijaBot & Tifratin & Ideophagous & Sedrati & Rachidourkia \\ \cline{2-7}
 & {\textbf{Count}} & \makecell{3,285 (34\%)} & \makecell{1,302 (13.47\%)}& \makecell{1,231 (12.74\%)}& \makecell{765 (7.92\%)} & \makecell{540 (5.59\%)} \\ \cline{2-7}
 & {\textbf{Type}} & Bot & Human & Human & Human & Human \\
\hline
\end{tabular}}
\caption{\label{tab:3}Top five page creators in the Arabic Wikipedia editions, highlighting their types (bots or humans) and how many articles they have created until March 1st, 2024, according to Wikiscan Statistics service.\protect\footnotemark}
\end{table*}

\footnotetext{Wikiscan Statistics service: \href{https://wikiscan.org}{https://wikiscan.org}.}

\subsubsection{Type of Contributors}
\label{sec:2.4.2}
\begin{figure*}[!htp]
    \centering
    \includegraphics[width=\linewidth]{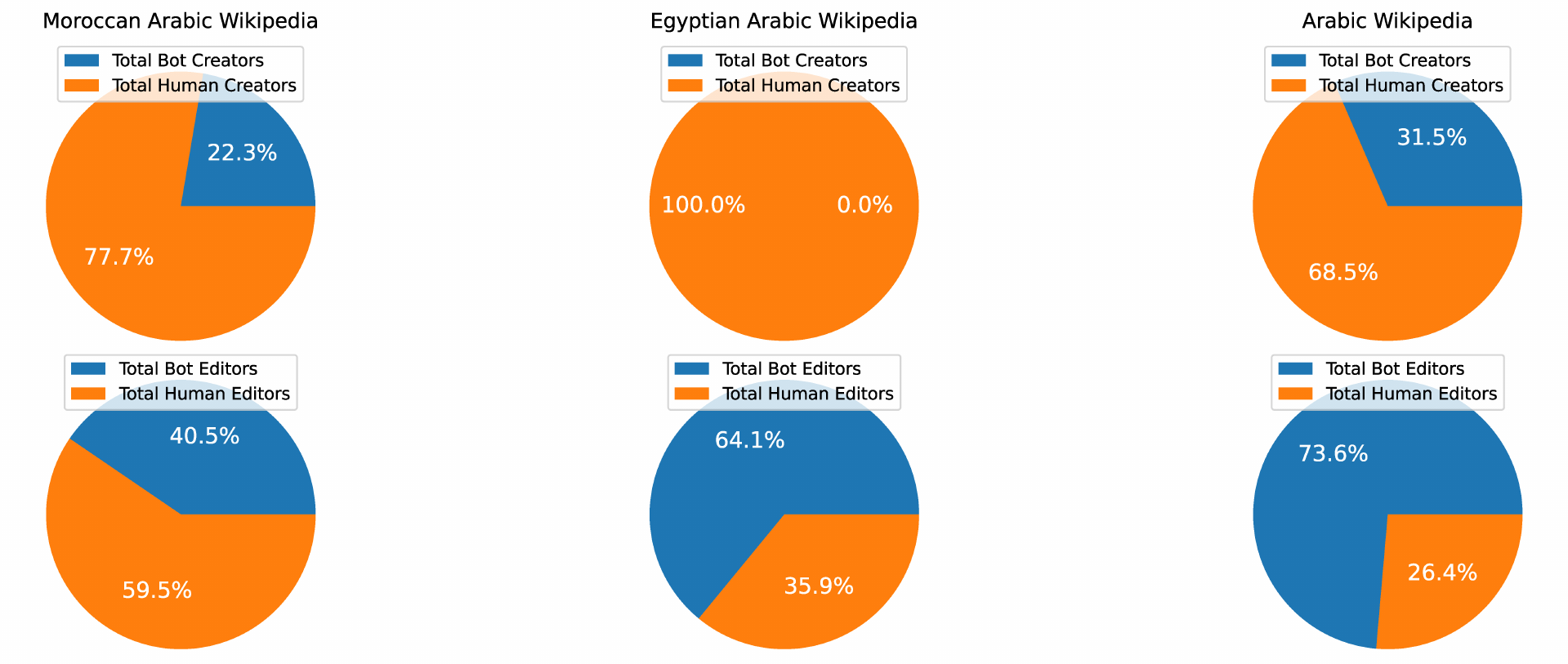}
    \caption{Visualizations displaying the percentage of article creators and editors in terms of their types, bots, and humans, and their number of contributions (article creations) in each Arabic Wikipedia edition.}
    \label{fig:3}
\end{figure*}

We calculate the percentage of creators and editors of articles (bots and humans) in each Arabic Wikipedia edition. We use the absolute count of page creators and classify the creators based on their types, bots or humans, while with the page editors, we calculate the percentage using the total number of editors on each article and set a threshold of 50\%, where if an article was edited by more than 50\% by bots, we then consider this article a bot-edited, and vice versa. As shown in Figure \ref{fig:3}, we see bots often create articles side-by-side with humans in the Arabic Wikipedia (31.5\%) and Moroccan Wikipedia (22.30\%) editions, which is normal and permitted to a certain degree according to Wikipedia’s bot policy \citep{25}. However, in the Egyptian Wikipedia edition, we observe that its articles are 100\% created by humans, i.e., registered users, and this percentage is misleading given that 42.72\% of its articles are automatically template-translated from English to Arabic using templates without human supervision or intervention, as documented by \citet{1} and  \citet{11}.

\section{Experimental Setup}
\label{sec:3}
We, here, attempt to build classifiers to identify and mitigate the impacts of the template-translated articles on the Egyptian Wikipedia edition since it particularly suffers from template translations, as documented by  \citet{11}. We first extract all articles with their metadata, split the articles into two categories: before and after the template-based translation occurred, and lastly, label, preprocess, and encode these categorized articles using Arabic pre-trained models.

\subsection{Dataset Filtrating and Labeling}
\label{sec:3.1}
We follow a few heuristic rules to classify Egyptian Wikipedia into articles created before and after the massive template-based translation activities related to creation dates, total edits, and types of creators and editors. We take insights from our exploratory analysis, section \ref{sec:2}, the Wikimedia Statistics service, and the previous research works that documented the template translation activities in the Egyptian Wikipedia \citep{1, 11, 113}, to craft these rules, specifically when selecting the dates.

\begin{table}[h!]
\centering
\tiny\renewcommand{\arraystretch}{1}
\resizebox{\linewidth}{!} {
\begin{tabular}{|c|c|} 
\hline
 \textbf{Category} &  \textbf{Total} \\
\hline
 \textbf{Total Articles (both categories)} & 736,107 \\
\hline
\textbf{Articles Before Template Translation} & 11,126 \\
\hline
\textbf{Articles After Template Translation} & 155,275 \\
\hline
\textbf{Uncategorized Articles} & 569,706 \\
\hline
\end{tabular}}
\caption{\label{tab:4}Statistics of filtered articles after applying our heuristic filtration rules, displaying the totals.}
\end{table}

We list the heuristic rules for filtering the articles created \emph{before} and \emph{after} the translations in Appendix \ref{app:b}, where we employ more rigorous heuristic rules to filter the articles created \emph{after} the template translation appeared on the Egyptian Wikipedia. In Table \ref{tab:4}, we show the statistics of our rule-based filtration process. We then randomly select 10K articles from each category to train a multivariate machine learning classifier to detect the template-based translations automatically. We lastly label the articles \emph{before} translation as ‘human-generated’ articles since all articles are created by registered users and label the articles \emph{after} translation as ‘template-translated’ articles.



\subsection{Dataset Preprocessing}
\label{sec:3.2}
We lightly preprocess the filtered articles by replacing all non-alphanumeric and non-Arabic characters with white spaces and normalizing the extra unnecessary whitespaces to one whitespace. We do not apply stemming, lemmatization, or any Arabic text normalization on the articles to have organic content (articles) as much as possible.

\subsection{Dataset Encoding}
\label{sec:3.3}
We use two different types of embedding techniques to encode the randomly selected 20K articles separately: pre-trained Egyptian Arabic context-independent word embeddings (Word2Vec) of the size of 300 dimensions from \texttt{Spark-NLP} Python library\footnote{Word2Vec Embeddings in Egyptian Arabic (300d): \href{https://sparknlp.org/2022/03/14/w2v\_cc\_300d\_arz\_3\_0.html}{https://sparknlp.org/2022/03/14/w2v\_cc\_300d\_arz\_3\_0}.} and context-dependent word embeddings (contextual) of the size of 768 dimensions produced by utilizing the pre-trained CAMeLBERT-Mix POS-EGY model\footnote{CAMeLBERT-Mix POS-EGY model: \href{https://huggingface.co/CAMeL-Lab/bert-base-arabic-camelbert-mix-pos-egy}{https://huggingface.co/CAMeL-Lab/bert-base-arabic-camelbert-mix-pos-egy}.} \citep{52} as our feature extraction model. The goal is to test with different embedding techniques to maximize the performance of our multivariate machine learning classifiers and investigate how the type and size of the word embeddings would affect their performance.

\section{Template Translation Detection}
\label{sec:4}
We experiment with a few supervised classification algorithms and unsupervised clustering algorithms to determine which approach and algorithm will best solve our template-based translation problem.

\subsection{Input Features Extraction}
\label{sec:4.1}
\begin{figure*}[!htp]
    \centering
    \includegraphics[width=\linewidth]{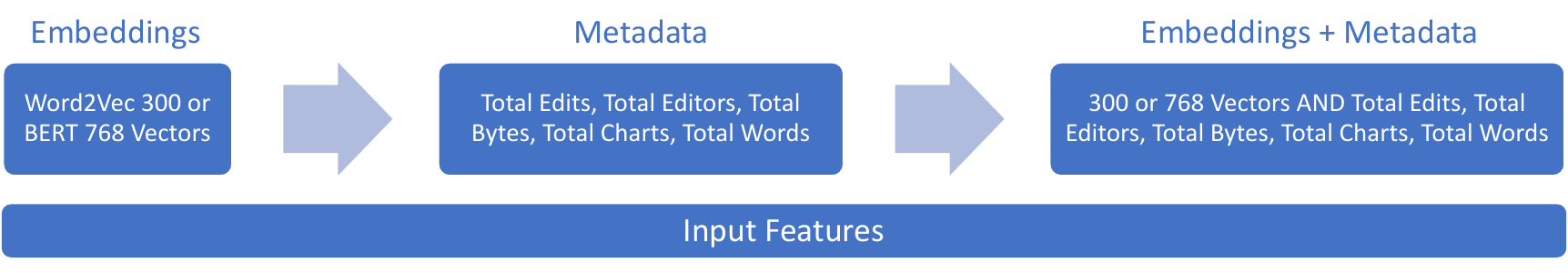}
    \caption{A basic process chart demonstrating the studied input features: embeddings (two word embeddings of sizes 300 or 768), metadata (five metadata of articles), or both (embeddings + metadata).}
    \label{fig:4}
\end{figure*}

We aim to leverage the metadata of corpus, i.e., articles, collected using Wikimedia services to detect the template-translated articles in the Egyptian Wikipedia edition. Besides utilizing pre-trained Word2Vec and CAMeLBERT word embeddings as input features, we also include the metadata we collect about every article: total edits, total editors, total bytes, total characters (charts), and total words. Overall, we test the machine learning algorithms' performance using three input features: only embeddings, only metadata, or both (metadata and embeddings), as illustrated in Figure \ref{fig:4}.

\subsection{Metadata Ablation Studies}
\label{sec:4.2}
We perform two ablation studies for each machine learning algorithm (classification and clustering) to determine the best metadata features to include in the input features. We first test each metadata feature’s performance individually and then combine two, three, and all metadata features consecutively.

\subsection{Classification Algorithms}
\label{sec:4.3}
We select five supervised classification algorithms to solve our multivariate classification problem: Logistic Regression (LR) \citep{53}, Support Vector Machine (SVM) \citep{54}, Gaussian Naive Bayes (GNB) \citep{55}, Random Forests (RF) \citep{56}, and XGBoost (eXtreme Gradient Boosting) \citep{57}. We, in the next subsections, discuss the experimental setups and the performance results of these supervised machine learning classifiers.

\subsubsection{Classification Experimental Setup}
\label{sec:4.3.1}
We split the randomly selected 20K articles into training (80\%) and testing (20\%) splits with data shuffling and stratification enabled to ensure that the training and test splits are randomized and have the same proportion of each class. We further evaluate our classifiers using the accuracy metric with the Stratified K-Folds Cross-Validation technique, where we set the number of folds K=5, ensuring every fold has a representative class distribution.

\subsubsection{Results of Classification Ablations}
\label{sec:4.3.2}
We report, in Table \ref{tab:5}, the evaluation accuracy results on the testing splits of our metadata ablations. We can see that all machine learning classifiers achieve excellent (100\%) to very good performance (100\%>accuracy>90\%) with the total edits and total editors separated or combined. In contrast, metadata features like the total bytes, total characters, and total words perform from fairly to poorly and, unfortunately, decrease the overall performance of all metadata features combined with some classifiers like SVM. Generally, we observe that the ensemble classifiers (RF and XGBoost) outperform the other classifiers even with the metadata features that contribute less to the classifiers’ learning.

\begin{table*}[h!]
\centering
\tiny\renewcommand{\arraystretch}{1}
\resizebox{\textwidth}{!} {
\begin{tabular}{|c|c|c|c|c|c|c|c|c|} 
\hline
\multirow{2}{*}{\textbf{Classifier}} & \multicolumn{8}{c|}{\textbf{Metadata}} \\
 \cline{2-9}
  & {\textbf{A}} & {\textbf{B}} & {\textbf{C}} & {\textbf{D}} & {\textbf{E}} & {\textbf{A+B}} & {\textbf{C+D+E}} & {\textbf{All}} \\
\hline
{\makecell {\textbf{Logistic} \textbf{Regression}}} & 100 & 100 & 88.30 & 83.85 & 84.67 & 100& 89.03 & 98.42 \\
\hline
{\makecell {\textbf{Support} \textbf{Vector} \textbf{Machine}}} & 90.30 & 100 & 87.95 &83.60 & 83.95 & 99.78 & 87.62 & 87.75 \\
\hline
{\makecell {\textbf{Naive} \textbf{Bayes}}}& 100 & 100 &82.00 &74.28 & 78.00 & 100 &  80.50 & 99.60 \\
\hline
{\makecell {\textbf{Random} \textbf{Forest}}}& 100 & 100 &86.17 & 82.23 & 84.80 &  100 &  91.25 & 100 \\
\hline
\textbf{XGBoost} & 100 & 100 & 88.60 & 84.52 & 84.70 & 100 & 90.53 & 100 \\
\hline
\end{tabular}}
\caption{\label{tab:5}Accuracies of metadata ablations of the studied classifiers. Encoded columns denote metadata features as follows: A) total edits, B) total editors, C) total bytes, D) total characters, and E) total words.}
\end{table*}

\begin{table*}[h!]
\centering
\tiny\renewcommand{\arraystretch}{1}
\resizebox{\textwidth}{!} {
\begin{tabular}{|c|c|c|c|c|c|} 
\hline
\multirow{2}{*}{\textbf{Classifier}} & \multicolumn{2}{c|}{\textbf{Embeddings}} & \multirow{2}{*}{\textbf{Metadata}} & \multicolumn{2}{c|}{\textbf{Both (Embeddings + Metadata)}}\\
 \cline{2-3}
\cline{5-6}
&\textbf{Word2Vec} & \textbf{CAMeLBERT} & &\textbf{Word2Vec} & \textbf{CAMeLBERT} \\
\hline
{\makecell {\textbf{Logistic} \textbf{Regression}}} &91.22 & 99.30 & 98.42 & 99.40 & 100  \\
\hline
{\makecell {\textbf{Support} \textbf{Vector} \textbf{Machine}}} & 99.02 & 98.45 &87.75 & 87.90 & 87.90  \\
\hline
{\makecell {\textbf{Naive} \textbf{Bayes}}}& 88.90 & 95.17 & 99.60 & 99.60 & 99.52  \\
\hline
{\makecell {\textbf{Random} \textbf{Forest}}} & 98.08 &98.17 & 100 & 100 & 99.95  \\
\hline
\textbf{XGBoost} &98.28 & 98.78 & 100& 100 &100  \\
\hline
\end{tabular}}
\caption{\label{tab:6}Accuracies of the machine learning classifiers studied, showing their performance with different input features: two embedding styles, corpus metadata, and both embeddings and metadata combined.}
\end{table*}

\subsubsection{Results of Classification Algorithms}
\label{sec:4.3.3}
We show, in Table \ref{tab:6},   the evaluation accuracy scores on the testing splits of the multivariate machine learning classifiers studied, demonstrating how the classifiers would perform with three input features: two embedding styles (Word2Vec or CAMeLBERT), corpus/articles metadata, and both embeddings and metadata combined. Here, we decided to include all the articles' metadata, not only the features that performed well in our ablation studies, to diversify the classifiers’ learning and ensure that each category of the Egyptian Wikipedia articles (human-generated and template-translated) is well-represented. We report, again, that the SVM classification algorithm underperforms all the other algorithms and find that the metadata features present a bottleneck performance for it (i.e., highly variable features). We attribute the poor performance to the complex, multivariate nature of the dataset, specifically, the high variability of the metadata features like the total bytes, words, and characters, as seen in Table \ref{tab:5}.\footnote{We handled the dataset noise through our filtration process and the bias by balancing the dataset classes, yet the dataset variance is challenging due to the high dispersion in metadata features collected. } On the other bright side, we find that ensemble classification algorithms like RF and XGBoost excel and outperform the traditional, single classification algorithms due to their ability to overcome noise, bias, and variance; the RF algorithm uses the bagging technique, and XGBoost algorithm uses boosting technique to handle such technical challenges.
\footnote{As an online application, we deploy our best classifier, XGBoost, with input features of metadata and CAMeLBERT embeddings. See Appendix \ref{app:c} for details.}


\subsection{Clustering Algorithms}
\label{sec:4.4}
We explore three different unsupervised clustering algorithms to solve the template-based translation problem: K-Means \citep{58}, Hierarchical Agglomerative \citep{59}, and DBSCAN (Density-Based Spatial Clustering of Applications with Noise) \citep{60}. We, in the following, discuss the experimental setups and the performance results of these unsupervised machine learning clusterers.

\subsubsection{Clustering Experimental Setup}
\label{sec:4.4.1}
We feed the unsupervised clustering algorithms all the randomly selected 20K articles after removing the labels without splitting them due to their nature. We set the number of clusters to K=2 since our dataset only has two categories (human-generated and template-translated). We further evaluate our clusterers using the Silhouette coefficient with the Euclidean distance, a widely used internal evaluation metric to measure how cohesive and separated the clusters are, based on the distances or similarities between the data points, i.g., articles.\footnote{Values of the Silhouette coefficient are always between 1 and -1. We apply a percentage normalization (multiply values by 100) when reporting the values to draw a head-to-head comparison between algorithms.}

\subsubsection{Results of Clustering Ablations}
\label{sec:4.4.2}
We report the Silhouette scores of our metadata ablation studies in Table \ref{tab:7}. We can see that all machine learning clusterers achieve great performance with the total bytes, total characters, and total words, separated or combined, except for the DBSCAN algorithm. In contrast, metadata features like the total edits and total editors perform from fairly to poorly with K-Means and Hierarchical clustering algorithms, except for the DBSCAN algorithm. The results of these metadata ablations indicate an opposite behavior from those discussed in subsection \ref{sec:4.3.2}, where the previously weak metadata features for the classification algorithms, like the total bytes, words, and characters, became strong metadata features for the clustering algorithms instead of the total edits and editors, which were previously strong. Generally, the K-Means and Hierarchical clustering algorithms outperform the DBSCAN algorithm even with the metadata features that contribute more to the clusterers’ learning.

\begin{table*}[!htp]
\centering
\tiny\renewcommand{\arraystretch}{1}
\resizebox{\textwidth}{!} {
\begin{tabular}{|c|c|c|c|c|c|c|c|c|} 
\hline
\multirow{2}{*}{\textbf{Clusterer}} & \multicolumn{8}{c|}{\textbf{Metadata}} \\
 \cline{2-9}
&  {\textbf{A}} & {\textbf{B}} & {\textbf{C}} & {\textbf{D}} & {\textbf{E}} & {\textbf{A+B}} & {\textbf{C+D+E}} & {\textbf{All}} \\
\hline
\textbf{K-Means} & 82.68 & 78.32 & 97.10 & 96.46 & 96.39 & 81.77 & 96.89 & 96.89 \\
\hline
\textbf{Hierarchical} & 86.85 & 81.42 & 97.10 & 97.37 & 97.32 & 82.28 & 96.08 & 97.52 \\
\hline
\textbf{DBSCAN} & 97.80 & 99.62 & 37.20 & 67.58 & 89.79 & 77.11 & 68.35 & 68.33 \\
\hline
\end{tabular}}
\caption{\label{tab:7}Silhouette scores of the metadata ablations of the studied clusterers.  Encoded columns denote metadata features: A) total edits, B) total editors, C) total bytes, D) total characters, and E) total words.}
\end{table*}

\begin{table*}[!htp]
\centering
\tiny\renewcommand{\arraystretch}{1}
\resizebox{\textwidth}{!} {
\begin{tabular}{|c|c|c|c|c|c|} 
\hline
\multirow{2}{*}{\textbf{Clusterer}} & \multicolumn{2}{c|}{\textbf{Embeddings}} & \multirow{2}{*}{\textbf{Metadata}} & \multicolumn{2}{c|}{\textbf{Both (Embeddings + Metadata)}}\\
 \cline{2-3}
\cline{5-6}
&\textbf{Word2Vec} & \textbf{CAMeLBERT} & &\textbf{Word2Vec} & \textbf{CAMeLBERT} \\
\hline
\textbf{K-Means} & 12.50 & 14.95 & 96.89 & 96.89 & 96.89 \\
\hline
\textbf{Hierarchica} & 11.79 & 10.82 & 97.52 & 96.77 & 96.77 \\
\hline
\textbf{DBSCAN} & 61.64 & 8.43 & 68.33 & 68.34 & 68.68 \\
\hline
\end{tabular}}
\caption{\label{tab:8}Silhouette scores of the machine learning clusterers studied, showing their performance with different features: two embedding styles, corpus/articles metadata, and both embeddings and metadata.}
\end{table*}

\subsubsection{Results of Clustering Algorithms}
\label{sec:4.4.3}
We show, in Table \ref{tab:8}, the Silhouette scores of the machine learning clusterers studied, demonstrating how the unsupervised clusterers would perform with three input features: two embedding styles (Word2Vec or CAMeLBERT), corpus/articles metadata, and both embeddings and metadata combined. We, here, fit all the articles’ metadata, not only the features that performed well in our ablation studies, to diversify the clusterers’ learning and ensure that each class of the Egyptian Arabic Wikipedia articles (human-generated and template-translated) is included. We report that all the clustering algorithms perform poorly with the word embeddings as features, whereas the metadata features present a performance improvement. We assume clustering the word embeddings is challenging, especially with their large dimensionality; Word2Vec’s size is 300, and CAMeLBERT’s is 768. 
Overall, the unsupervised clustering algorithms underperform the supervised classification algorithms, yet we can confirm that the clustering algorithms do better with low-dimensionality features like articles’ metadata, even though they introduce high-variable and dispersed features.

\section{Discussion}
\label{sec:5}
We discuss three negative implications of the unguided, unreviewed, unsupervised template-based translation from English to Arabic on the Egyptian Wikipedia articles: societal, representation, and performance implications. On the societal implications, we argue that using off-the-shelf-translation tools like Google Translate, which is widely known for its social problems like gender, cultural, and religious biases and stereotypes, could not only cause linguistic and grammatical errors but also amplify these social risks like biases and stereotypes \citep{65, 66, 67, 68, 9}. Many researchers have emphasized how unsupervised translations are prone to serious gender bias issues, like producing translations with inaccurate gender, that could impact native speakers. For example, \citet{69} have automatically evaluated the gender bias for eight highly-gendered languages like Arabic and found that a few popular industrial and academic machine translation systems (like Google Translate and Microsoft Translator\footnote{Microsoft Bing: \href{https://www.bing.com/translator}{https://www.bing.com/translator}.}) were significantly prone to gender-biased translation errors for all tested target languages. We believe those machine translation systems are greatly beneficial tools, yet they should not be used to naively, directly, or automatically translate content without human review, especially if the content is related to the societal representation of Arabic native speakers.

On the representation implications, we argue that such automatic template-based translations without humans in the loop could misrepresent the Egyptian Arabic native speakers, where instead of the Egyptian people enriching the content of Wikipedia by sharing their voices, opinions, knowledge, perspectives, and experiences, a couple of registered users automated the creation and translation of more than a million and a half million articles (95.56\%) from English on their behalf without supervision or revision of the translated articles, disregarding that the main goal of Wikipedia is to be written by the people to the people \citep{6}.  Another troubling drawback of such a practice is the cultural misrepresentation of the Egyptian people and their culture, where the unfiltered and unsupervised translation from English could introduce content that is not representative of the culture of native speakers. Lastly, we argue that including culturally unrepresentative articles from the Egyptian Arabic Wikipedia in pre-training corpora for language models could present cultural implications and generate culturally misaligned outputs from these models, where the majority of Arabic and multilingual language models have been fundamentally pre-trained on Wikipedia dumps like Jais and Jais-chat \citep{62}, AraMUS \citep{63}, and JASMINE \citep{64}. We believe research works, like ours, that automatically identify these template-translated articles could promote data transparency and help researchers make an informed decision about what to include in their pre-training datasets/corpora. 

On the performance implications, we argue that the template-based translations that occurred on the Egyptian Wikipedia produce not only short and shallow articles, where we have reported that nearly 46\% of the Egyptian Wikipedia articles are less than 50 tokens/words and recognized a large number of duplicate n-grams due to the templates used in translations, but also articles that lack lexical richness and diversity, where we have found that the Egyptian Wikipedia scored the worst among other Arabic Wikipedia editions in the MTLD metric. These poorly translated articles could negatively impact the performance of language models and NLP tasks that are trained on them. One research that supports our claim is the recent work of \citet{13}, where they documented that models trained on the template-translated articles of the Egyptian Wikipedia performed the worst when compared with the models trained on the Arabic Wikipedia articles. 
Finally, we recommend excluding the unfiltered template-translated articles from Egyptian Wikipedia from training datasets to mitigate their negative societal, representation, and performance implications and encourage using automatic detection systems, like ours, to identify such articles that are not only mispicturing the Egyptian people and their culture but also affecting the performance of language models and NLP tasks.

\vspace{-3pt}
\section{Limitations}
\label{sec:6}
We leverage five metadata of articles of different sizes (total edits, total editors, total bytes, total characters, and total words) and then append them to two types of word embeddings (Word2Vec and CAMeLBERT) of sizes of 300 or 768 vectors to build powerful classifiers, yet concatenating all these different features could produce highly variable features due to the high dispersion between the extracted input features, which could present a performance challenge for our proposed automatic detection system and could increase the non-deterministic behavior of its classifiers.

\vspace{-3pt}
\section{Conclusion}
\label{sec:7}
We attempt to mitigate the template translations on the Egyptian Arabic Wikipedia by identifying these template-translated articles and their characteristics through exploratory analysis and developing automatic detection systems. We first investigate the content of the three Arabic Wikipedia editions in terms of density, quality, and human contributions and use such insights to build powerful multivariate machine learning classifiers leveraging articles’ metadata to detect template-translated articles automatically; we find that the supervised classification algorithms are better than the unsupervised clustering algorithms. We then publicly deploy the best-performing classifier, XGBoost, as an application and release the extracted, filtered, labeled, and preprocessed datasets to the community to benefit from our datasets and the online detection system.

\vspace{-3pt}
\section*{Reproducibility}
We share our labeled datasets, code and scripts of the exploratory analysis, and the multivariate machine learning classifiers on GitHub at \hspace{1pt}{\fbseries \url{https://github.com/SaiedAlshahrani/leveraging-corpus-metadata}}.

\section*{Acknowledgments}
We would like to thank Clarkson University and the Office of Information Technology (OIT) for providing computational resources. We also would like to thank Norah Alshahrani for her valuable feedback.

\newpage
\nocite{*}
\section*{Bibliographical References}
\vspace{-20pt}
\bibliographystyle{lrec-coling2024-natbib}
\bibliography{lrec-coling2024-example}


\appendix

\section{Analysis of N-Grams}
\label{app:a}
We analyze the 5-grams and 10-grams closely since they are suitable, not long or short. The n-grams in the Egyptian Wikipedia are very large compared to the Arabic and Moroccan Wikipedia editions, as indicated in Tables \ref{tab:9} and \ref{tab:10}. Plus, it is noticeable that these counts do not decay exponentially as they normally should (the larger the n-gram size, the smaller the n-grams’ count) but linearly and slowly (all near 222K even with different sizes of n-grams), suggesting this abnormal decay is a symptom of the template translations that Egyptian Wikipedia suffered from, where some grams/parts/phrases from the used templates are frequently and constantly repeated. 

We additionally observe that most of the top ten 5-grams and 10-grams of the Moroccan Wikipedia edition are predominantly non-Arabic grams, which seems in a format of the Wikitext Markup Language \cite{71}, as exhibited in Tables \ref{tab:9}, \ref{tab:10}, and \ref{tab:11}. We further investigate this issue by testing our parsing code scripts and find that it does not occur when parsing articles from the other two Arabic Wikipedia editions, Arabic (AR) and Egyptian (ARZ), using the same code scripts; it only surfaces when parsing the Moroccan Wikipedia articles. We attribute this issue to either leaking Wikipedia templates used to create articles or insert images into articles or an issue with the method used to dump and compress Moroccan Wikipedia articles. We urge the global and local admins of the Moroccan Wikipedia edition to investigate this issue, which could affect not only the Moroccan Wikipedia content but also the performance of perspective NLP models and tasks trained on such content.

\begin{table*}[!htp]
\centering
\begin{tabular}{|c|c|c|} 
\hline
\textbf{Wikipedia} & \textbf{Count} & \textbf{5-Gram} \\
\hline
\multirow{3}{*}{} & 141,880 & \makecell{\<تصنيف أشخاص على قيد الحياة>\\Classification of surviving people}\\

\cline{2-3}
 \textbf{Arabic (AR)} & 80,460 & \makecell{\<لاعبو كرة قدم رجالية مغتربون>\\Men’s football players expatriate}\\

\cline{2-3}
 & 38,793 & \makecell{\<تعداد عام وبلغ عدد الأسر>\\A general census, and the number of families reached}\\
 
 \hline
\multirow{3}{*}{} & 222,964 & \makecell{\<صوره هيا مجال الكره السماويه>\\It is a picture of the celestial sphere}\\

\cline{2-3}
 \textbf{Egyptian (ARZ)} & 222,961 & \makecell{\<الكره السماويه اللى المجره جزء>\\The celestial sphere, of which the galaxy is a part}\\

\cline{2-3}
 & 222,939 & \makecell{\<مجموعه من النجوم اللى بتكون>\\A collection of stars that forms}\\
 
  \hline
\multirow{3}{*}{} & 5,172 & \makecell{\\ width text textcolor black fontsize \\\textcolor{white}{.}}\\

\cline{2-3}
 \textbf{Moroccan (ARY)} & 2,057 & \makecell{\<لعاطلين اللي سبق ليهوم خدمو>\\For unemployed people, who have previously served}\\

\cline{2-3}
 & 1,483 & \makecell{\<على حساب لإحصاء الرسمي عام>\\According to the official census of the year}\\
 
\hline
\end{tabular}
\caption{\label{tab:9} Selected top three 5-grams from each Arabic Wikipedia edition with their English translations.}
\end{table*}

\begin{table*}[!htp]
\centering
\tiny\renewcommand{\arraystretch}{1}
\resizebox{\textwidth}{!} {
\begin{tabular}{|c|c|c|} 
\hline
\textbf{Wikipedia} & \textbf{Count} & \textbf{10-Gram} \\
\hline
\multirow{3}{*}{} & 38,790 & \makecell{\<تعداد عام وبلغ عدد الأسر أسرة وعدد العائلات عائلة مقيمة>\\
A general census, the number of families was one family, and the number of families was one resident family}\\

\cline{2-3}
 \textbf{Arabic (AR)} & 38,710 & \makecell{\<وبلغت نسبة الأزواج القاطنين مع بعضهم البعض من أصل المجموع>\\
 The percentage of couples living together was out of the total}\\

\cline{2-3}
 & 38,524 & \makecell{\<نسبة منها لديها أطفال تحت سن الثامنة عشر تعيش معهم>\\
 A percentage of them have children under the age of eighteen living with them}\\
 
 \hline
\multirow{3}{*}{} & 222,935 & \makecell{\<صوره هيا مجال الكره السماويه اللى المجره جزء منها الانزياح>\\
A picture of the celestial sphere, of which the galaxy is a part of the displacement}\\

\cline{2-3}
 \textbf{Egyptian (ARZ)} & 222,935 & \makecell{\<المطلع المستقيم هوا الزاويه المحصوره بين الدايره الساعيه لجرم سماوى>\\
The right ascension is the angle enclosed between the hourly circle of a celestial body}\\

\cline{2-3}
 & 221,251 & \makecell{\<أو صوره هيا مجال الكره السماويه اللى المجره جزء منها>\\
 Or a picture of the celestial sphere, of which the galaxy is a part}\\
 
  \hline
\multirow{3}{*}{} & 2,586 & \makecell{\\ imagesize width height plotarea left right top bottom timeaxis orientation \\
\textcolor{white}{.}}\\

\cline{2-3}
 \textbf{Moroccan (ARY)} & 1,483 & \makecell{\<لعاداد كان ديالو واصل شخص على حساب لإحصاء الرسمي عام>\\
 The number of people was counted up to according to the official census of the year}\\

\cline{2-3}
 & 1,348 & \makecell{\<ما كايعرفوش يقراو ولا يكتبو نسبة كان قاريين فوق انوي>\\
They did not know how to read or write the percentage of literate was above}\\
 
\hline
\end{tabular}}
\caption{\label{tab:10} Selected top three 10-grams from each Arabic Wikipedia edition with their English translations.}
\end{table*}

\begin{table*}[!htp]
\centering
\tiny\renewcommand{\arraystretch}{1}
\resizebox{\textwidth}{!} {
\begin{tabular}{|r|} 
\hline
\<توريرت (سيدي أحمد وعبدالله): أرمد هو دوار مجمع كاين جماعة أسني دائرة أسني إقليم لحوز جهة مراكش آسفي لمغريب هاد وار كينتامي مشيخة إمليل لعاداد> \\
\< كان ديالو واصل شخص على حساب لإحصاء الرسمي عام هو دوار لي كاين الجبل السلسلة ديال لأطلس الكبير الغربي الجغرافيا دوار أرمد بعيد كلم على مدينة>\\
\< مراكش على ارتفاع حوالي ميترو على البحر فالجبال ديال الأطلس الكبير هاد الدوار مشهور بلفلاحة خصوصا التفاح والڭرڭاع فيه بزاف لوبيرجات والمحلات ولعشاش>\\
\< دالتجارة السياحية والبيع دالمنتجات التقليدية والماكلة السكان إحصائيات عامة عدد السكان ديال أرمد تزاد عدد لفاميلات تزاد مابين عدد لبالغين كان واحد منهوم دكور >\\
imagesize width height plotarea left right top bottom timeaxis orientation vertical alignbars justify colors id gray value gray dateformat yyyy \<نتوات> \\
period from till scalemajor unit year increment start gridcolor gray plotdata bar color green width from till width text textcolor black fontsize px \\
bar color red width from till width text textcolor black fontsize px imagesize width height plotarea left right top bottom timeaxis orientation vertical \\
alignbars justify colors id gray value gray dateformat yyyy period from till scalemajor unit year increment start gridcolor gray plotdata bar color \\
 
\<لفاميلات الجواج لولادة نسبة > width from till width text textcolor black fontsize px bar color red width from till width text textcolor black fontsize px green \\
\<الجواج أرمد واصلة لموعد ال لعمر عند الجواج اللولاني هو عام عند الرجال عند لعيالات لخصوبة عند لعيالات واصلة لخصوبة لكاملة التسكويل نسبة التسكويل واصلة > \\
\<نسبة لأمية واصلة لخدمة نسبة الناس النشيطين دوار أرمد واصلة نسبة الشوماج واصلة نوطات عيون لكلام تصنيف جهة مراكش آسفي تصنيف دوار لمغريب تصنيف دوار>\\
\< إقليم لحوز تصنيف مقالات فيها مصدر بايت.>\\ 

\hline
\end{tabular}}
\caption{\label{tab:11} A sample of a parsed article from Moroccan Wikipedia, showing the embedded Wiki markups.}
\end{table*}

\section{Heuristic Filtration Rules}
\label{app:b}
\begin{figure}[!htp]
    \centering
    \includegraphics[width=\linewidth]{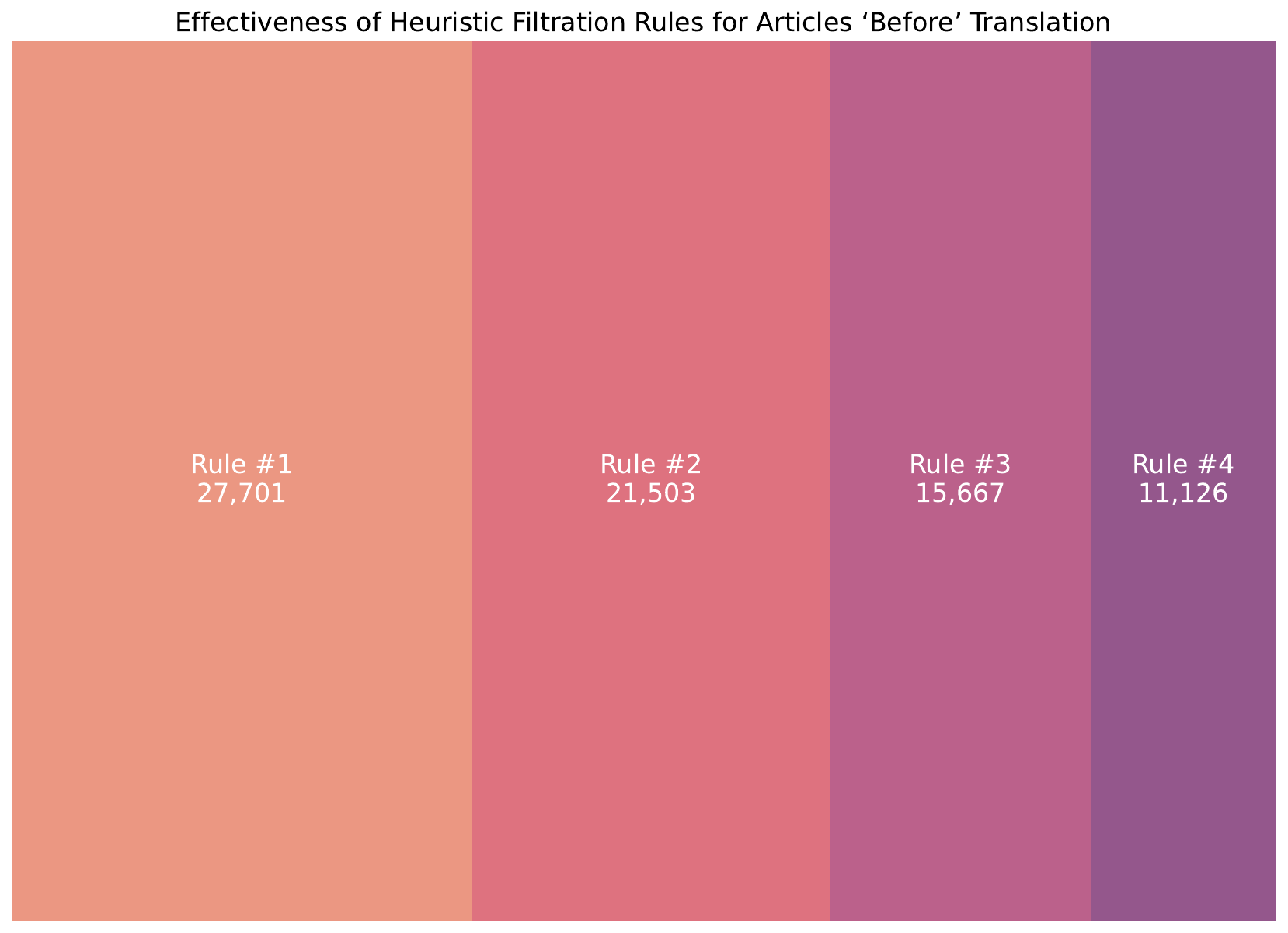}
    \caption{A treemap showing the effectiveness of the heuristic rules for articles \emph{before} the template-based translation in Egyptian Wikipedia, highlighting the number of articles filtered out by each rule.}
    \label{fig:5}
\end{figure}

\begin{figure}[!htp]
    \centering
    \includegraphics[width=\linewidth]{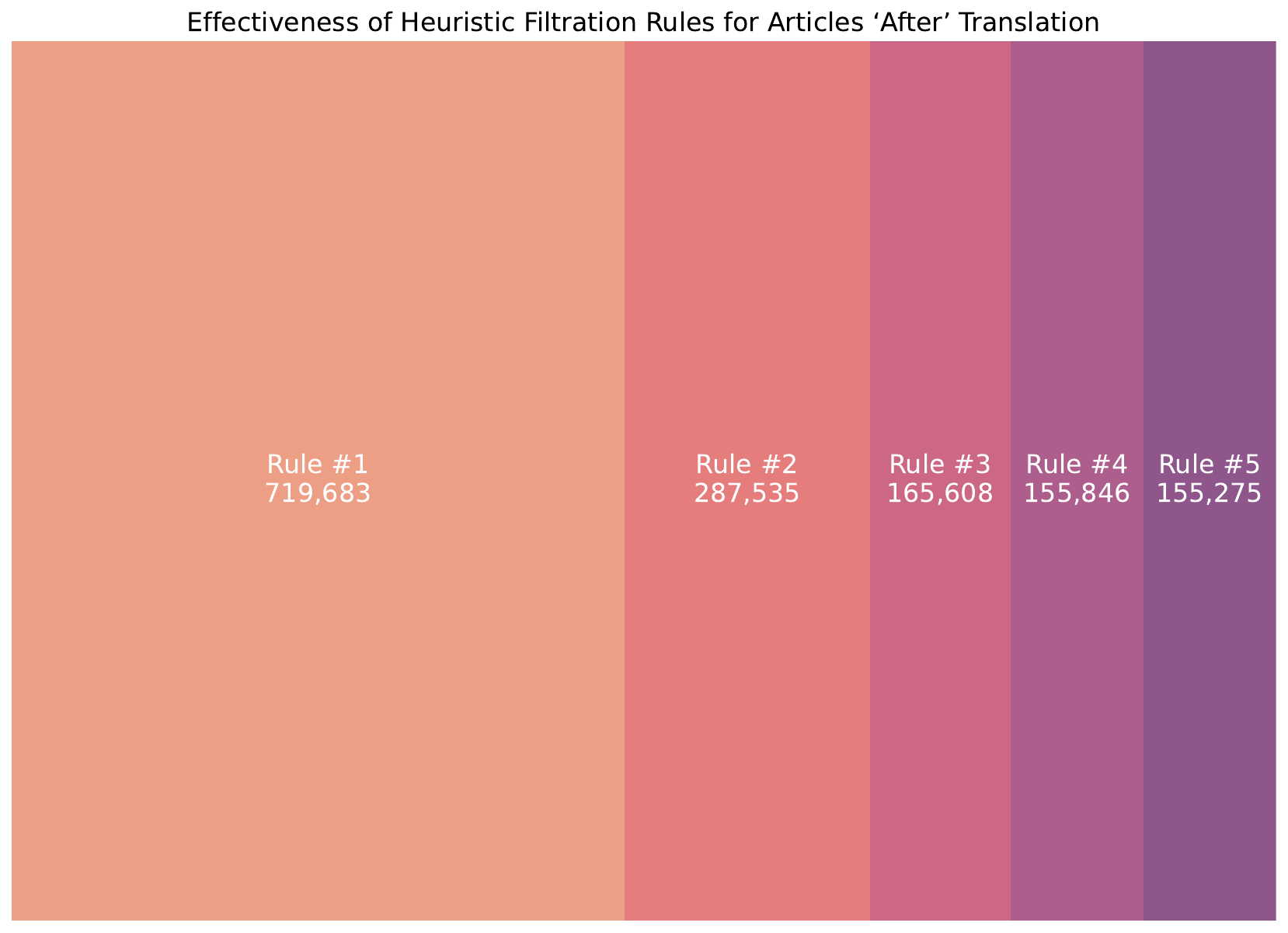}
    \caption{A treemap showing the effectiveness of the heuristic rules for articles \emph{after} the template-based translation in Egyptian Wikipedia, highlighting the number of articles filtered out by each rule.}
    \label{fig:6}
\end{figure}

We list the heuristic filtration rules used to filter the articles \emph{before} and \emph{after} the template-based translation in the Egyptian Wikipedia edition and further shed light on the effectiveness of each enforced rule. We demonstrate, in Figures \ref{fig:5} and \ref{fig:6},  the effectiveness of the implemented rule-based filtration. We can see that our heuristic filtration rules are practical, as each rule consecutively and rigorously filters out unfit articles that do not meet the heuristic filtration rules.

\vspace{6pt}

\noindent$*$ \underline{Heuristic filtration rules for  \emph{before} the translation:}
\begin{enumerate}
  \item Include articles created before 2019-12-01.
  \item Include articles with more than five edits.
  \item Include articles with more than three editors.
  \item Include articles with greater than or equal to 50\% human editors.
\end{enumerate}

\noindent$*$ \underline{Heuristic filtration rules for \emph{after} the translation:}
\begin{enumerate}
  \item Include articles created between 2019-12-1 and 2023-12-01 and discard young articles with an age of less than 30 days (2023-12-1 and 2024-1-1).
  \item Include articles with less than five edits.
  \item Include articles with less than three editors.
  \item Include articles with greater than or equal to 50\% bot editors.
  \item Include articles created by these registered users, ‘HitomiAkane’ and ‘Al-Dandoon’, who overwhelmed the Egyptian Arabic Wikipedia with massive auto-generated and template-translated articles without human supervision.
\end{enumerate}

\section{\textsc{Egyptian Wikipedia Scanner}}
\label{app:c}
\begin{figure}[!htp]
    \centering
    \includegraphics[width=\linewidth]{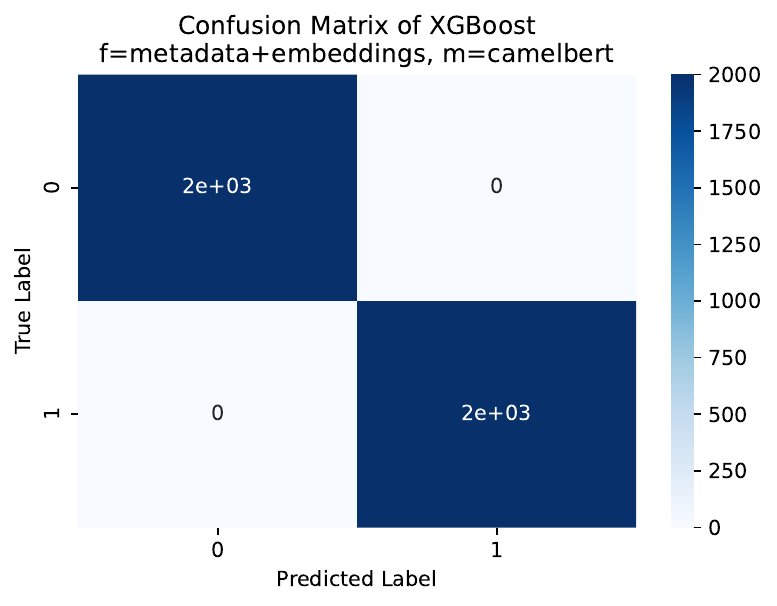}    

    \caption{Confusion matrix of the best, deployed classifier, XGBoost, which takes input features of articles’ metadata combined with CAMeLBERT’s embeddings, showing the excellent performance of this multivariate ensemble classifier.}
    \label{fig:7}
\end{figure}

\begin{figure}[!htp]
    \centering

    \includegraphics[width=\linewidth]{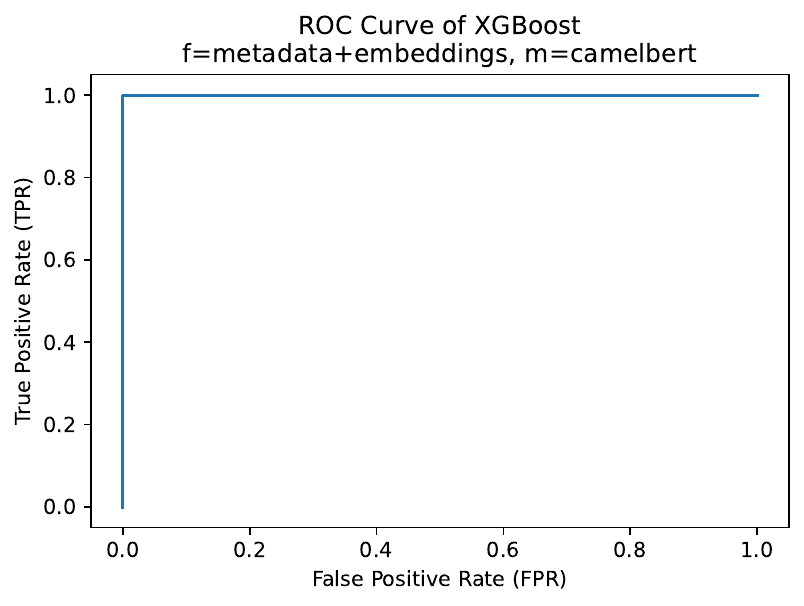}
    \caption{ROC curve of the best, deployed classifier, XGBoost, which takes input features of articles’ metadata combined with CAMeLBERT’s embeddings, showing the excellent performance of this multivariate ensemble classifier.}
    \label{fig:8}
\end{figure}

We evaluate our multivariate supervised machine learning classifiers using metrics like accuracy and ROC-AUC (Receiver Operating Characteristic Area Under Curve). We then publicly deploy and host our best classifier, XGBoost, which takes input features of articles’ metadata and CAMeLBERT embeddings, as illustrated in Figures \ref{fig:7} and \ref{fig:8}. We include the articles’ metadata because we find that, from our two ablation studies, metadata could be practical and encode features useful for the classifier’s learning. We also choose CAMeLBERT over Word2Vec word embeddings because CAMeLBERT’s embeddings take the context into account, and  Word2Vec’s embeddings are context-free and need to be retrieved word by word and then averaged for the whole article; this is not ideal.

\begin{figure*}[!htp]
    \centering
    \vspace{10pt}
    \includegraphics[width=\linewidth]{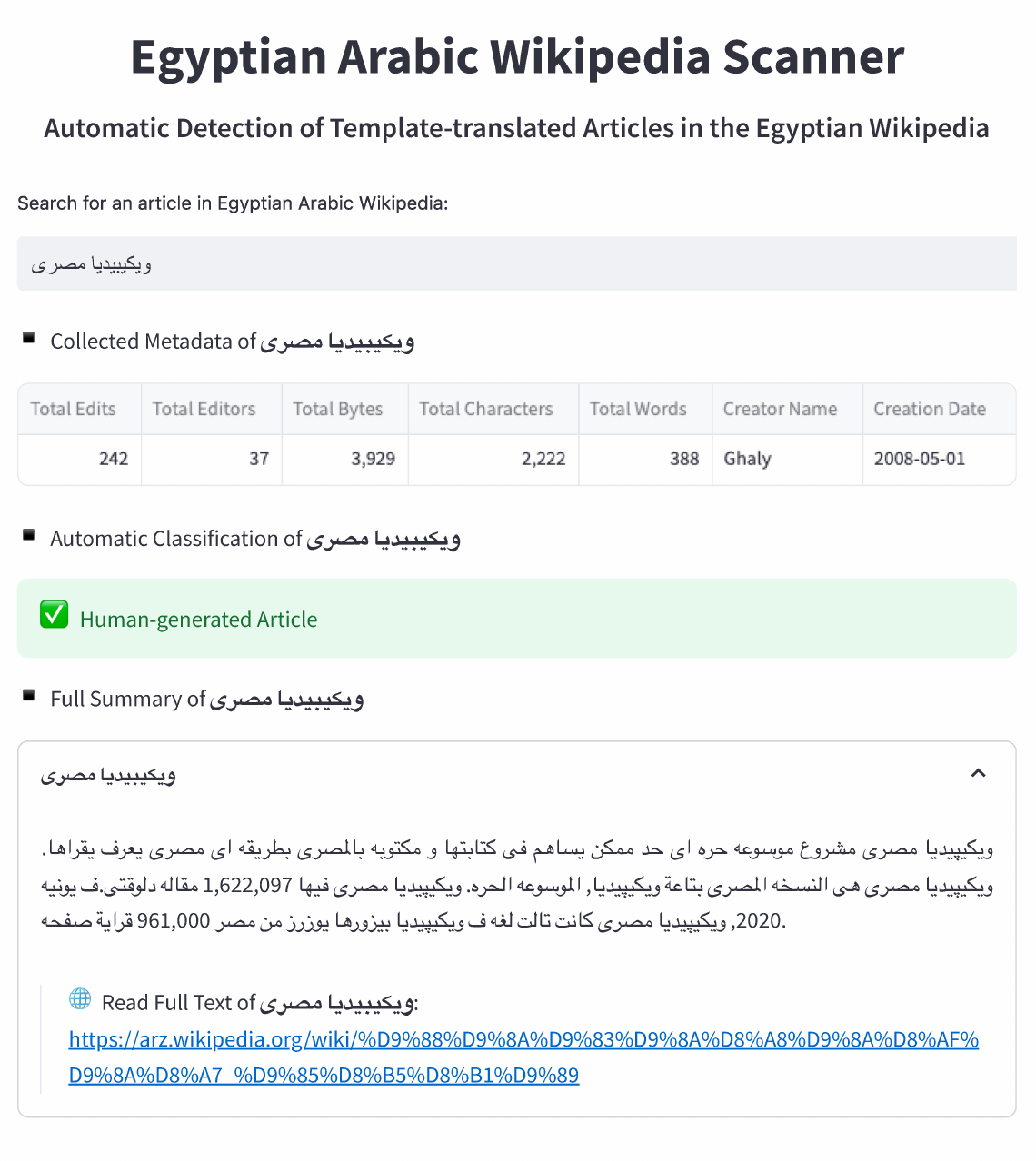}   

    \caption{A screenshot of the \textsc{Egyptian Wikipedia Scanner}, illustrating its capabilities and features.}
    \label{fig:9}
\end{figure*}

We call this online application \textsc{Egyptian Wikipedia Scanner}, where users can search for an article directly or select a suggested article retrieved using fuzzy search from the Egyptian Arabic Wikipedia edition.
The application automatically fetches the article’s metadata (using the Wikimedia \texttt{XTools} API), displays the fetched metadata in a table, and automatically classifies the article as `human-generated' or `template-translated'. The application also dynamically displays the full summary of the article and provides the URL to the article to read the full text, as shown in Figure \ref{fig:9}.

We utilize the Streamlit Framework\footnote{Streamlit Framework: \href{https://streamlit.io}{https://streamlit.io}.} to design, host, and deploy the application on the free Streamlit Community Cloud\footnote{Streamlit Cloud: \href{https://streamlit.io/cloud}{https://streamlit.io/cloud}.} service, making it publicly accessible to everyone at \href{https://egyptian-wikipedia-scanner.streamlit.app}{https://egyptian-wikipedia-scanner.streamlit.app}. We also host the application on Hugging Face Spaces to avoid running out of Streamlit Cloud free, limited resources: \href{https://huggingface.co/spaces/SaiedAlshahrani/Egyptian-Wikipedia-Scanner}{https://huggingface.co/spaces/SaiedAlshahrani/Eg-yptian-Wikipedia-Scanner}. This online application, \textsc{Egyptian Wikipedia Scanner}, is open-sourced on GitHub with an MIT license, here: \href{https://github.com/SaiedAlshahrani/Egyptian-Wikipedia-Scanner}{https://github.com/SaiedAlshahrani/Egyptian-Wik-ipedia-Scanner}.

\vspace{26pt}

\vspace{25pt}

\end{document}